\documentclass[10pt,twocolumn,letterpaper]{article}

\usepackage[pagenumbers]{iccv} %

\definecolor{iccvblue}{rgb}{0.21,0.49,0.74}
\usepackage[pagebackref,breaklinks,colorlinks,allcolors=iccvblue]{hyperref}

\AtBeginDocument{%
  \providecommand\BibTeX{{%
    \normalfont B\kern-0.5em{\scshape i\kern-0.25em b}\kern-0.8em\TeX}}}

\usepackage{listings}
\usepackage{xcolor}
\usepackage{float}
\usepackage{graphicx}
\usepackage{amsmath}
\usepackage{booktabs}

\usepackage{tabu}
\usepackage{multirow}
\usepackage{rotating} %
\usepackage{caption}
\usepackage{subcaption}
\usepackage{pifont}
\newcommand{\cmark}{\checkmark}%
\newcommand{\xmark}{$\times$}%

\usepackage{soul}
\usepackage{svg}

\DeclareMathAlphabet{\altmathcal}{OMS}{cmsy}{m}{n}
\DeclareMathAlphabet{\mathbfit}{OT1}{ptm}{bx}{it}

\newlength\paramargin
\newlength\figmargin

\newlength\secmargin
\newlength\figcapmargin
\newlength\tabcapmargin

\long\def\ignorethis#1{}

\definecolor{methodred}{RGB}{248, 203, 173}
\definecolor{methodyellow}{RGB}{245, 220, 143}
\definecolor{methodgreen}{RGB}{197, 224, 180}
\definecolor{methodblue}{RGB}{180, 199, 231}

\newcommand{\tb}[1]{\textbf{#1}}

\newbox\jsavebox%

\makeatletter
\newcommand{\providelength}[1]{%
  \@ifundefined{\expandafter\@gobble\string#1}
   {%
    \typeout{\string\providelength: making new length \string#1}%
    \newlength{#1}%
   }
   {%
    \sdaau@checkforlength{#1}%
   }%
}

\newcommand{\sdaau@checkforlength}[1]{%
  \edef\sdaau@temp{\expandafter\sdaau@getfive\meaning#1TTTTT$}%
  \ifx\sdaau@temp\sdaau@skipstring
    \typeout{\string\providelength: \string#1 already a length}%
  \else
    \@latex@error
      {\string#1 illegal in \string\providelength}
      {\string#1 is defined, but not with \string\newlength}%
  \fi
}
\def\sdaau@getfive#1#2#3#4#5#6${#1#2#3#4#5}
\edef\sdaau@skipstring{\string\skip}
\makeatother

\def\xi{\mathbf{x}_i}

\graphicspath{{figures}, {examples}}

\title{\name: Consistent Keyframe Synthesis for Cinematic Scene Composition}

\author{
Quynh Phung$^{1\ast}$ \quad Long Mai$^{2}$ \quad Fabian David Caba Heilbron$^{2}$ \\
Feng Liu$^{2}$ \quad Jia-Bin Huang$^{1}$ \quad Cusuh Ham$^{2}$ \\
\\
$^1$~University of Maryland, College Park \quad
$^2$~Adobe Research \\
{\tt\small \{quynhpt,jbhuang\}@umd.edu} \quad
{\tt\small \{malong, caba, fengl, ham\}@adobe.com}\\
\url{https://cinevers.github.io/}
}

\newcommand{\Sref}[1]{Sec.~\ref{#1}}

\newcommand{\Fref}[1]{Fig.~\ref{#1}}
\newcommand{\Tref}[1]{Table~\ref{#1}}

\newcommand{\name}[0]{CineVerse}
\begin{document}

\twocolumn[{%
  \renewcommand\twocolumn[1][]{#1}%
  \maketitle
  \vspace{-3em}
  \begin{center}
    \captionsetup{type=figure}
    \includegraphics[width=\textwidth]{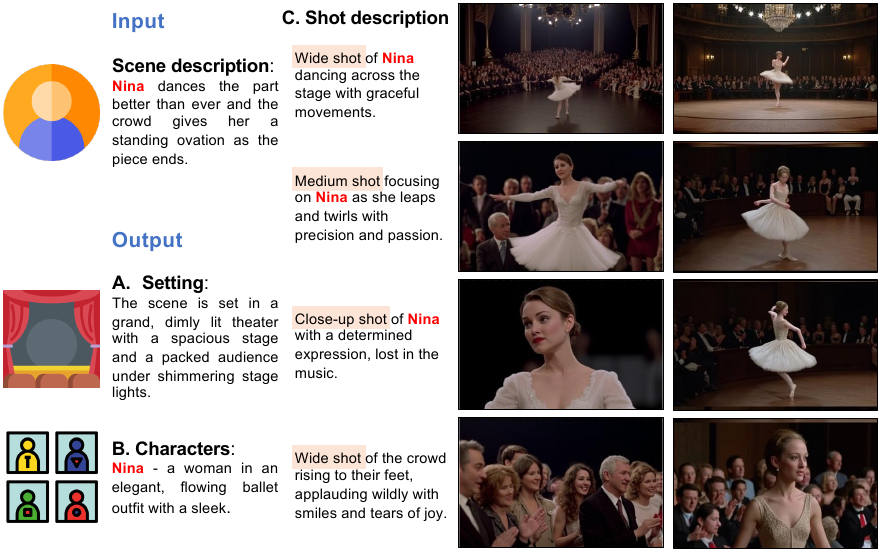}
    \begin{tabu} to \textwidth {X[2.3,c] X[1,c] X[1,c]}
      & \footnotesize Ours & \footnotesize IC‑LoRA~\cite{ic-lora} \\
    \end{tabu}
    \captionof{figure}{
      \textbf{Cinematic scene composition.} 
      Given a simple scene description, we prompt a pre‑trained language model to generate the setting, characters with unique appearances, and detailed shot descriptions with explicit shot sizes. 
      We then use this detailed scene plan to synthesize consistent keyframes using our fine‑tuned text‑to‑image model
      adapted from IC‑LoRA~\cite{ic-lora} specifically for our cinematic scene composition task.
      Compared to the baseline IC‑LoRA, our results showcase improved text‑image alignment, consistency, and continuity.
    }
    \label{fig:teaser}
  \end{center}
}]

\renewcommand{\thefootnote}{$\ast$}
\footnotetext{Work was done while interns at Adobe}

\begin{abstract}
We present \name, a novel framework for the task of cinematic scene composition. Similar to traditional multi-shot generation, our task emphasizes the need for consistency and continuity across frames. However, our task also focuses on addressing challenges inherent to filmmaking, such as multiple characters, complex interactions, and visual cinematic effects. In order to learn to generate such content, we first create the \name~dataset. We use this dataset to train our proposed two-stage approach. First, we prompt a large language model (LLM) with task-specific instructions to take in a high-level scene description and generate a detailed plan for the overall setting and characters, as well as the individual shots. Then, we fine-tune a text-to-image generation model to synthesize high-quality visual keyframes. Experimental results demonstrate that \name~yields promising improvements in generating visually coherent and contextually rich movie scenes, paving the way for further exploration in cinematic video synthesis.

\end{abstract}

\section{Introduction}

\label{sec:introduction}

In recent years, Gen AI technologies have significantly transformed the landscape of visual content creation. Advances in generative image and video models now make it possible to generate images and videos with impressive levels of realism. However, the generation of content that effectively supports visual storytelling remains an open problem. Current methodologies excel at producing individual images or video shots, or at generating loosely connected content that follows a storyline, but they often fall short in creating coherent sequences that reflect the structured nature of professional filmmaking.

\begin{figure}[t]
  \centering
    \includegraphics[width=1.0\columnwidth]{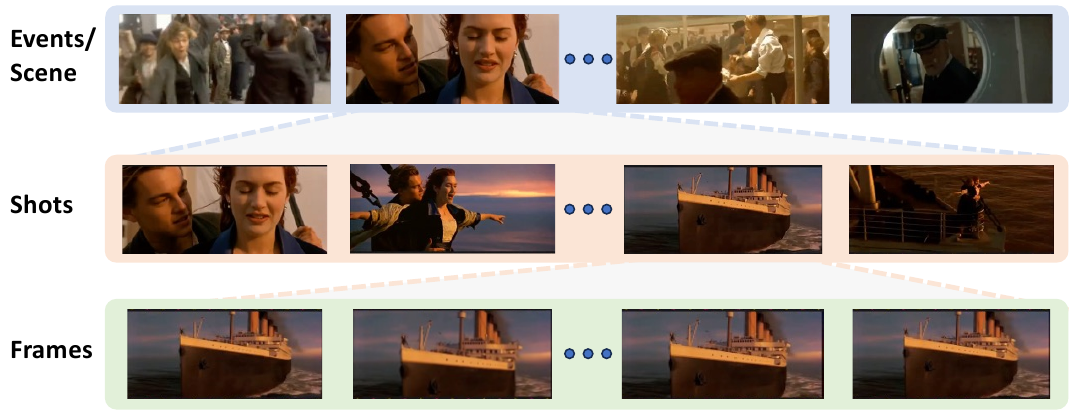}
  \caption{
  \tb{Movie structure.}
A movie is composed of unique scenes and events that drive the storyline. Each scene consists of multiple shots establishing context, highlighting character emotions, or emphasizing key details. At the finest level, individual frames bring these shots to life. 
Our work aims to empower everyday users to composite cinematic scenes at the shot level.
}
  \label{fig:structure}
\end{figure}

Professional filmmaking is characterized by a highly organized approach to storytelling. At the top level, a movie is composed of key events, often referred to as movie scenes. Each movie scene consists of a sequence of carefully crafted shots, which consist of a series of frames. The movie structure is depicted in ~\Fref{fig:structure}. These shots are assembled to capture the audience's attention and convey the director's creative intent. The sequences of shots often adhere to well-recognized structures that can be understood as the ``grammar'' of cinematic rules~\cite{arijon1976grammar}. For instance, a scene featuring an intense conversation between two characters typically begins with a \textit{wide} shot that establishes the characters within their environment. This is often followed by \textit{over-the-shoulder} shots that highlight the connection and engagement between the characters during their dialogue, culminating in a series of \textit{close-up} shots that capture each character's expressions as they take turns speaking.

A movie is composed of unique scenes and events that drive the storyline. Each scene consists of multiple shots establishing context, highlighting character emotions, or emphasizing key details. At the finest level, individual frames bring these shots to life. 
Our work aims to empower everyday users to composite cinematic scenes at the shot level

\begin{figure}[t]
    \vspace{-1.5em}
    \begin{subfigure}[b]{0.49\textwidth}
        \centering
        \includegraphics[width=1\textwidth]{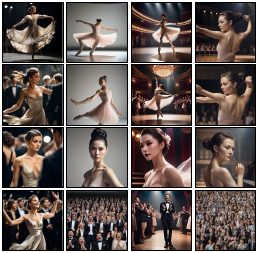}
        \vskip -0.2cm
        \begin{tabu} to \textwidth {X[1,c]X[1,c]X[1,c]X[1,c]}
        \footnotesize{1P1S~\cite{1prompt}} & \footnotesize{ConsiStory}~\cite{consistory} & \footnotesize{StoryDiff~\cite{storydiff}}& \footnotesize{VideoStudio~\cite{videodrafter}}
    \end{tabu}
    \end{subfigure}

    \vspace{-0.2cm}
\caption{
\tb{Limitations of existing work.} 
Existing multi-image text-to-image generation models struggle with complex prompts that require precise scene composition. They often fail to adhere to specified camera shots (e.g., wide, medium, close-up) and lack subject and setting consistency and continuity.
}
    \label{fig:limitation_baseline}
    \vspace{-1em}
\end{figure}

The key challenges in cinematic scene composition include the need to create a scene plan that effectively expands upon the scene description in a cinematically-meaningful manner, as well as the requirement to generate a storyboard that visualizes this plan. The storyboard must consist of a set of keyframes that are consistent with one another, both in terms of character representation and scene setup, while closely adhering to the established scene plan.

To tackle these challenges, we propose \name, a two-stage approach that leverages in-depth knowledge of filmmaking principles. The first stage employs an in-context prompting technique to guide a large language model (LLM) in generating a cinematically meaningful shot design. This method capitalizes on the LLM's strong semantic understanding, while also ensuring that the prompts connect this understanding to specific aspects of movie shots, such as shot types and their intended effects.

The resulting scene plan provides a sequence of shot descriptions that can be utilized as text prompts for text-to-image generation models in the second stage, facilitating the creation of a coherent storyboard. We further enhance this generation process by incorporating the recent In-Context LoRA (IC-LoRA) technique \cite{ic-lora}, which has demonstrated the ability to generate multiple images with high quality and consistent content. However, this method struggles to generate the correct number of shots. Our technical adjustments to IC-LoRA improve the model's capacity to generate the appropriate number of shots and enhance the fidelity of the generated images to the shot descriptions.

Other story-based generation methods (e.g., \cite{storydiff,consistory,1prompt}) face several common challenges, as shown in \Fref{fig:limitation_baseline}. First, because they generate individual frames sequentially, the lack of context being shared across the frames often leads to disruptions to the overall continuity, detracting from a coherent visual narrative. Second, characters and background have very limited variation across images due to the model's inflexibility in balancing recontextualizing these aspects of the scene and preserving their appearance. Third, these methods struggle with multi-character generation, which limits the scope of the input scenes. In contrast, \name~is able to generate coherent shots that are diverse in the camera shot size, perspective, and character appearance, while maintaining fidelity to the user's intent and continuity across frames. Our contributions are as follows:

\begin{enumerate}
    \item \textbf{Dataset.} We create the \name~dataset for cinematic scene composition. We build upon Storyboard20K \cite{storyboard20k} and refine existing scene descriptions to clearly identify the main characters of a given frame. We also add new attributes (shot descriptions, shot size, setting, and character descriptions) to enable fine-grained control over the generation process.
    \item \textbf{Scene planning.} We develop a technique to guide an LLM for cinematic scene planning. The structured format of the shot design enables users to make modifications to the visual story before it gets generated by the text-to-image model.
    \item \textbf{Cinematic scene composition.} We use our newly created dataset to fine-tune FLUX \cite{flux} using a low-rank adapter-based (LoRA) \cite{hu2022lora} approach. We leverage a strong spatial prior in FLUX to improve the reliability of generating sequences of varying lengths while adhering to the text, and maintaining continuity and consistency in appearance of characters and setting.
\end{enumerate}

\vspace{-1pt}
\section{Related work}
\label{sec:relatedwork}

\textbf{Learning from movie data}. 
Movies have long been a valuable resource for studying visual storytelling, providing structured narratives that guide advancements in video understanding~\cite{hollywood, ava, lsmdc, movieqa}. 
While early works focus on large-scale datasets for character and global story comprehension, scene-level analysis (e.g., shot composition analysis, transition detection, and cinematographic style classification) have gained prominence. 
Works such as MovieShots~\cite{movieshots} and CineScale~\cite{cinescale} provide annotations for shot types and camera movements, enabling studies on shot framing and stylistic trends. Additionally, datasets like FullShots~\cite{fullshots} and Shot2Story~\cite{shot2story} have expanded coverage by offering more diverse shot compositions and structured narrative understanding. More recently, Storyboard20K~\cite{storyboard20k} refined scene segmentation in the MovieNet dataset and adding scene-level attributes such as bounding boxes for characters and objects.

Despite progress in scene-level understanding, scene-level generation remains under-explored. Prior studies focus on classifying existing shots rather than generating coherent sequences. Some initial attempts, such as Film Editing Patterns~\cite{editpattern}, encode cinematic rules for virtual environments, but large-scale data-driven shot generation is still lacking. Existing shot datasets are often either limited in scope (focusing on a single cinematographic feature) or lack the scale necessary for generative models. Our work addresses this gap by curating a comprehensive scene-level dataset with detailed shot annotations, enabling both analysis and generation of visually coherent cinematic sequences.

\noindent\textbf{Story visualization}. Story visualization, which generates sequential images to depict a narrative, has become a prominent area in generative modeling. Early approaches, such as StoryGAN ~\cite{storygan} , showcased the potential of this technique by creating coherent visual sequences. More recent methods, including StoryDALL-E ~\cite{storydall}, leverage transformer architectures to boost performance, while diffusion-based models and training-free techniques~\cite{storydiff, consistory, 1prompt} enhance visual quality and identity preservation through shared key-value mechanisms. However, previous work primarily focus on generating events from overall story synopses. In contrast, our approach targets cinematic scene composition, focusing on generating keyframes corresponding to multiple consistent shots within the same scene, which demands stricter continuity and consistency. Moreover, although recent studies~\cite{seedx, seed-story, show-o, moviedreamer} demonstrate the potential of using LLM backbones to generate long story sequences, they typically require massive datasets and extensive computational resources. In this work, we achieve scene-level story visualization using a small dataset and limited resources.

\vspace{-1pt}
\section{Method}
\begin{figure}[t]
    \centering
    \includegraphics[width=\linewidth]{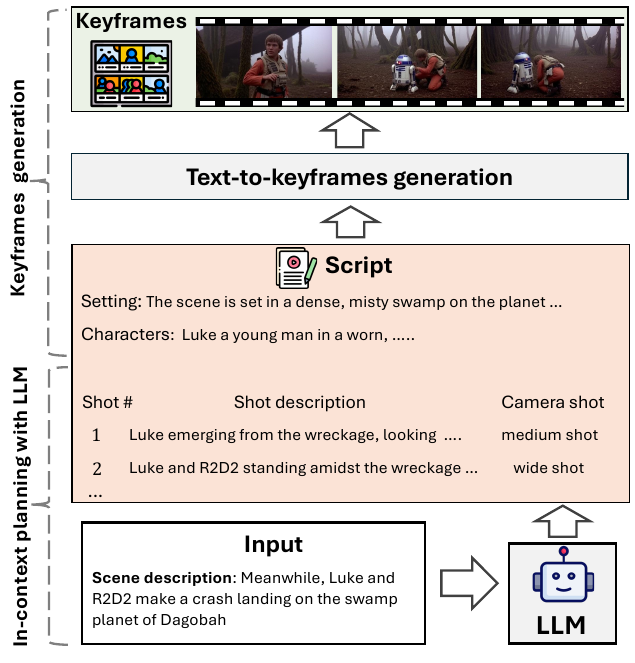}
    \caption{
\textbf{Method overview.}
In the stage 1, given the scene description as input, we leverage an LLM for in-context planning to produce a detailed \emph{script}.
This script consists of 
1) \emph{Setting}: A background description of the scene, 
2) \emph{Characters}: Individual characters with their unique appearances, and
3) \emph{Shot descriptions}: The context and actions of the characters along with specified camera shots.
In Stage 2, we use the generated script to synthesize multiple keyframes using text-to-image models fine-tuned on the proposed CineVerse dataset.
 }
    \label{fig:pipeline}
    \vspace{-1em}
\end{figure}

\name~takes as input a text description that summarizes the high-level content of a movie scene. Our goal is to generate storyboard keyframes that effectively visualize the narrative, capturing different camera angles and character focuses while maintaining consistency and continuity.

A main challenge in generating a coherent sequence of cinematic keyframes from a high-level scene description is that the model must not only understand the movie's narrative, but also reason and plan the sequence of shots. Expecting one model to handle both complex reasoning and generation tasks is unrealistic. Reflecting practical filmmaking, where detailed planning precedes shooting, our cinematic scene composition framework consists of two main stages: 1) \textit{In-context scene planning}, which defines the setting, characters, and descriptions for a sequence of shots, and 2) \textit{Keyframe generation}, which generates a sequence of keyframes. An overview of this process is shown in ~\Fref{fig:pipeline}.

\vspace{-0.1pt}
\subsection{In-context scene planning}

The primary objective of the scene planning step is to systematically convert textual scene descriptions into detailed shot-by-shot plans that guide subsequent keyframe generation. A robust scene plan captures essential cinematic elements, including framing, character positioning, camera shot size, and continuity across shots. Achieving clarity, coherence, and consistency in these planned descriptions is crucial, as it directly impacts the quality and coherence of the generated keyframes.

LLMs are recognized for their exceptional semantic comprehension, contextual reasoning, and narrative coherence capabilities. These attributes make LLMs particularly suited for the intricate task of shot planning, where understanding scene semantics, narrative progression, and visual storytelling nuances are essential. By leveraging these sophisticated reasoning abilities, \name~can generate more nuanced, contextually appropriate, and creatively detailed shot descriptions.

\noindent\textbf{Prompting LLMs for scene planning}. Generic LLM prompting, while powerful for general-purpose text generation, falls short when directly applied to shot planning tasks. For example, directly prompting an LLM, \textit{``Plan for the movie scene description: `Lucy jumps out of the train.'''}, the LLM responds, \textit{``Lucy leaps out of the speeding train, landing roughly and rolling on the ground. She quickly stands up, glancing nervously around, and then runs off towards a nearby wooded area.''}. Here, the LLM merely expands the description of the input scene, but does not provide additional details necessary to for filmmaking. Shot design in filmmaking demands specialized domain knowledge, including cinematic techniques, visual storytelling conventions, camera movements, and specific terminology unique to the filmmaking industry. 

To address these limitations, we develop a specialized prompting approach, explicitly designed for the cinematic domain. Our prompts integrate detailed cinematic knowledge, instructing the model explicitly on how to structure scene plans, or scripts. We utilize an instruction prompt to guide the LLM (illustrated in App. \ref{sec:gpt-4}). This prompt effectively enables the LLM to adapt to filmmaking by providing explicit guidance to generate scene plans with a precise format, including visual details, camera framing, and character attributes. Additionally, we provide exemplars after the instruction prompt to help the model better understand the task and follow the desired format, which can be found in \Fref{fig:prompt} in App. \ref{sec:appendix}.

\vspace{-0.1pt}
\subsection{Keyframe generation} \label{sec:keyframe-gen}

Given a detailed and structured script, we want to generate a sequence of consistent and coherent keyframes with cinematic elements. Inspired by IC-LoRA \cite{ic-lora}, \name~uses a single, detailed prompt to generate a set of images. The prompt contains special tokens to separate each frame and typically starts with a description of the overall sequence, followed by individual prompts for each shot. This structure helps the model capture both global context and the specific details of each frame.

IC-LoRA uses up to 100 examples to fine-tune FLUX \cite{flux}, a DiT \cite{peebles2023scalable}-based model, for the movie storyboard generation task using a set of low-rank adapters (LoRA) \cite{hu2022lora}. Unlike previous methods, IC-LoRA synthesizes multiple keyframes \textit{simultaneously} rather than \textit{sequentially/independently}. This method generates outputs that demonstrate a degree of temporal consistency and coherence between frames, closely resembling real movie sequences and demonstrating that the original FLUX model already possesses a prior for this task. 

However, there are a couple of key limitations of IC-LoRA for the cinematic scene composition task: 1) \textit{The mismatch between the number of planned shots for a given scene and the number of generated images}, and 2) \textit{Cropped content in the first or last frame}. The current approach uses a single concatenated text prompt, forcing the model to rely solely on the provided input image size to determine the number of frames. However, the dataset exhibits variable frame sizes across different scenes, making it difficult for the model to decide the correct number of generated frames, especially with large the number of images . Additionally, the existing pre-processing method sometimes crops parts of the first and last frames, resulting in incomplete images being generated (can be found in App. \Fref{fig:lora}).

Combining the observations of existing priors in the base FLUX model and the limitations of IC-LoRA, we explore reliably generating longer sequences of complete frames and enhancing textual control by adapting the IC-LoRA approach. Specifically, to resolve the first issue of mismatched expected and generated keyframes, we propose fixing the height of all frames to create a uniform dataset. We also add a distinct border between frames in the concatenated image to clearly indicate frame boundaries, providing an explicit spatial cue to the model for image separation. To resolve the cropped generated images, we eliminate the cropping step during pre-processing  to ensure that the entire content of the first and last frames is preserved. Lastly, to better enable our goal of generating cinematic images, we fine-tune our approach using a dataset of real movie keyframes paired with detailed description for each frame (described in Sec.~\ref{sec:dataset}).

\vspace{-1pt}
\section{\name~Dataset} \label{sec:dataset}
To facilitate cinematic scene composition, we introduce the \name~dataset, specifically designed for shot-level analysis. The \name~dataset provides detailed annotations for each image, capturing both narrative context and visual attributes. Unlike existing movie datasets, which often lack shot-level textual descriptions or focus solely on narrative content while omitting crucial visual details, our dataset offers a comprehensive and structured account of every shot. An overview of the information provided for each scene in our dataset can be seen in \Fref{fig:dataset}.

\vspace{-0.1pt}
\subsection{Data collection and filtering}

In cinematic scene composition, it is important to distinguish between traditional keyframes and multi-shot keyframes. In movie analysis, a keyframe is a single representative image selected from a specific shot or scene to capture its essential visual content. In contrast, multi-shot keyframes refer to a sequence of keyframes, each representing different shots within a scene. This multi-shot approach provides a more comprehensive overview of the narrative by capturing the continuity and variation within a scene, rather than relying on just one snapshot per scene.

We build upon the Storyboard20K \cite{storyboard20k} dataset, which contains nearly 20k scenes. However, we focus exclusively on 8.5k multi-shot scenes. The Storyboard20K dataset offers detailed annotations, including bounding boxes for characters and objects, as well as scene settings and descriptions. Additionally, we enrich our dataset by incorporating global reference information from the MovieNet  dataset \cite{movienet}, such as plot annotations, to further enhance the movie scene descriptions.

\vspace{-0.1pt}
\subsection{Refining existing attributes}

\begin{figure}[t]
    \centering
    \includegraphics[width=0.48\textwidth]{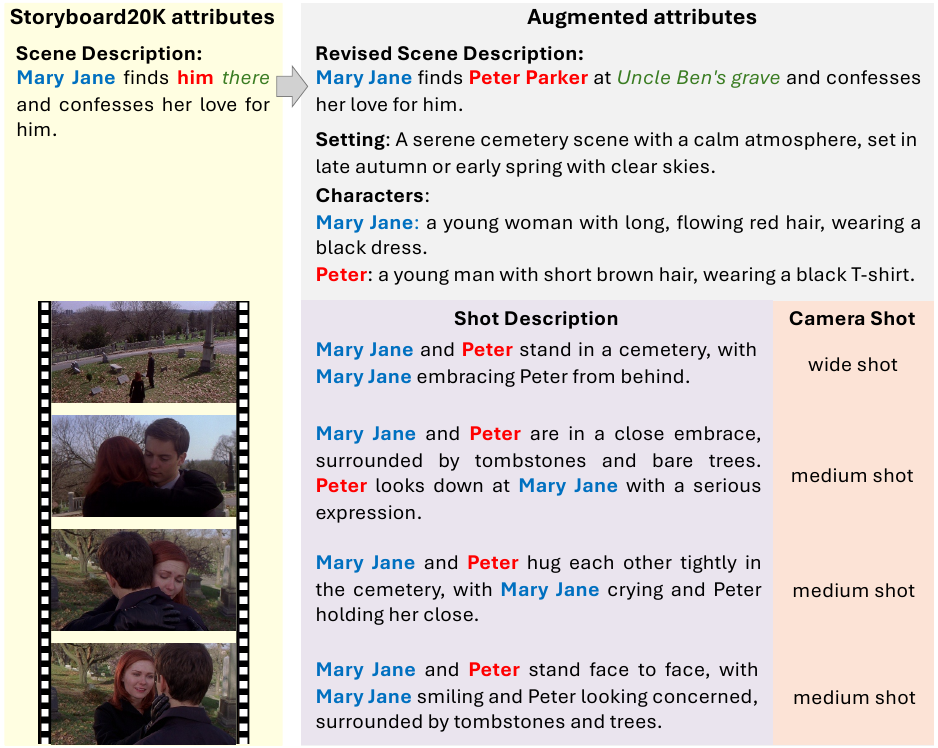}
    \caption{
    \textbf{Augumenting dataset}. 
    Storyboard20K~\cite{storyboard20k} scene descriptions are often ambiguous, including pronouns (e.g, him, there), making it difficult to produce keyframes with consistent scene and characters. 
    We augment the dataset by replacing the co-reference with a specific person/object/place. 
    We further extract detailed setting/character/shot descriptions and camera shot, forming a complete script to study cinematic scene composition. 
    }
    \label{fig:dataset}
    \vspace{-1em}
\end{figure}

We make small refinements to the existing \textit{scene descriptions} from Storyboard20K. The original scene descriptions in Storyboard20K often rely on pronouns and abbreviated character names. This practice can limit the ability to connect different scenes within a movie, compromising narrative continuity and making it difficult to maintain clear character identities. To address this issue, we combine the original scene descriptions with the global information in the plot (from the MovieNet dataset ~\cite{movienet}), which provides the full story context and helps clarify ambiguous references. For example, as shown in \Fref{fig:dataset}, the original description ``Mary Jane finds him there and confesses her love for him'' leaves it unclear who ``him'' or ``there'' refers to. We use an LLM, specifically LLama3.3 \cite{llama}, to analyze the plot and infer that ``him'' is ``Peter Parker'' and ``there'' is ``Uncle Ben's grave''. Then the refined scene description becomes ``Mary Jane Watson finds Peter Parker at Uncle Ben's grave and confesses her love for him,'' ensuring greater clarity and consistency throughout the narrative.

\vspace{-0.1pt}
\subsection{Extracting new attributes}

\begin{figure*}[t]
    \centering
    \vspace{-2em}
    \includegraphics[width=1.0\textwidth]{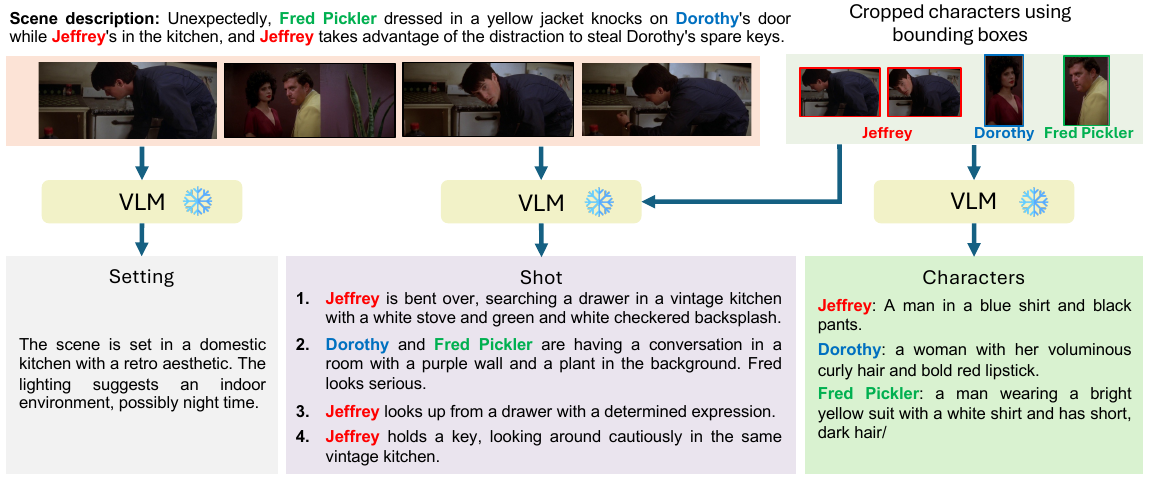}
    \caption{
    \textbf{Extract data attributes using LLaVa-OneVision}.
We use a pre-trained Vision-Language Model to extract the setting description, shot details, and character appearances.
    }
    \label{fig:llava}
    \vspace{-1em}
\end{figure*}

While storytelling analysis, for which Storyboard20K was designed for, shares some similarities with our specific scenario, it is not directly suited for generating multiple keyframes for individual movie scenes. To adapt the dataset for our task of cinematic shot composition, we leverage LLaVa-OneVision \cite{llavaone}, a multimodal LLM (MLLM), to extract \textit{shot descriptions}, \textit{shot size}, \textit{setting}, and \textit{character descriptions} (see \Fref{fig:llava}).

\noindent\textbf{Shot descriptions} are crucial in filmmaking as they clearly define visual elements, camera angles, character positions, and emotions, guiding filmmakers toward a unified creative vision. We input the keyframes from a scene as input to LlaVa-OneVision to generate a description for the contents of each image. To avoid generating shot descriptions with generic terms (e.g., ``man'', ``woman'') or pronouns, we also provide the model with a portrait of each character as context. We create the portraits by leveraging existing character name and bounding box annotations from Storyboard20K.

\noindent\textbf{Shot size} effectively guides viewers' attention and communicates emotion, narrative clarity, and visual storytelling in film. We annotate keyframes as \textit{close-up}, \textit{medium}, or \textit{wide} shots. Close-up shots highlight emotions and subtle details, while medium shots focus on character interactions, and wide shots establish the scene's context and environment. We use an off-the-shelf shot-size classification model ~\cite{shotsize} to automatically label each image.

\noindent\textbf{Setting}, including location, props, and weather, establishes the foundation for the theme and visual depiction of a scene. For instance, a nighttime scene with vibrant, colorful lighting can evoke a playful mood. We input the sequence of keyframes from a scene and prompt LLaVa-OneVision to generate scene-level descriptions for the setting.

\noindent\textbf{Character descriptions} include information about each character’s appearance, covering attributes such as costume, gender, and style. This attribute is particularly important for setting the tone of a scene (e.g., characters attending an elegant party should be depicted with beautiful dresses or suits to match the event's formality). For each character in a scene, we select up to three portraits (using the same process as in the shot description pipeline) as input to LlaVa-OneVision to generate scene-level character details. Using more images is costly, while fewer images might not capture all the character's details.

\vspace{-1pt}
\section{Experiments}
\label{sec:experiment}

In this section, we describe our experimental setup, baselines, and evaluation metrics. We assess \name~for the cinematic scene composition task using automated heuristics and user studies. First, we verify the ability of the scene planning stage to generate coherent and structured composition plans. Second, we examine whether the generation stage produces consistent, cinematic keyframes that align with the given plan and the original input scene description. 

\begin{table}[t!]
\centering
\small
\caption{ GPT‑4 preference percentages comparing our detailed prompts (with instruction and exemplars) against generic and instruction‑only prompts (0–100\%; higher is better)}
\label{tab:llm_gpt4}
\setlength{\tabcolsep}{2pt}
\begin{tabu} to 1.0\linewidth {@{}X[2.5,l]X[0.8,c]X[0.8,c]X[0.8,c]X[0.8,c]X[0.8,c]X[1.1,c]@{}}
\toprule
 \multirow{2}{*}{Method} & \multicolumn{2}{c}{Textual Align.} & \multicolumn{2}{c}{Consistency} & \multicolumn{2}{c}{Continuity} \\
\cmidrule(lr){2-3} \cmidrule(lr){4-5} \cmidrule(lr){6-7}

 & Scene & Shot  & Char  & BG  & Action & Camera  \\
\midrule
Generic prompt & 83.33 & 92.45 & 90.57 & 90.57 &  92.45 & 92.45  \\
Instruction &70.37 & 71.51 & 75.00 & 73.24 & 71.52 & 72.15\\

\bottomrule
\end{tabu}
\vspace{-1em}
\end{table}

\subsection{Experimental setup}

\textbf{Training details}. We randomly select 1000 scenes with 3-10 shots from \name~for training. Since the original Storyboard20K dataset is biased toward scenes with 3-4 shots, we filter out scenes to achieve a balanced training set. As described in \Sref{sec:keyframe-gen}, we also introduce a 16 px (the minimal value that can be independently encoded by FLUX) border with a distinct checkboard pattern between consecutive frames, and resize the images to a fixed height of 272 px (divisible by 8) while the width remains variable. We follow the IC-LoRA preprocessing approach of vertically concatenating the images within a scene.

We use LLama3.3-70B~\cite{llama}, the state-of-the-art open-source LLM, for the scene planning stage. The instruction prompt used to guide the LLM is shown in the supplementary document. For the keyframe generation stage, we fine-tune the base FLUX.1-dev model \cite{flux} using LoRA \cite{hu2022lora} with rank 128 and a learning rate of $1 \times 10^{-4}$ for approximately 16k steps.

\begin{figure*}[h]
    \begin{subfigure}[b]{1.\textwidth}
        \centering
        \includegraphics[width=1.\textwidth]{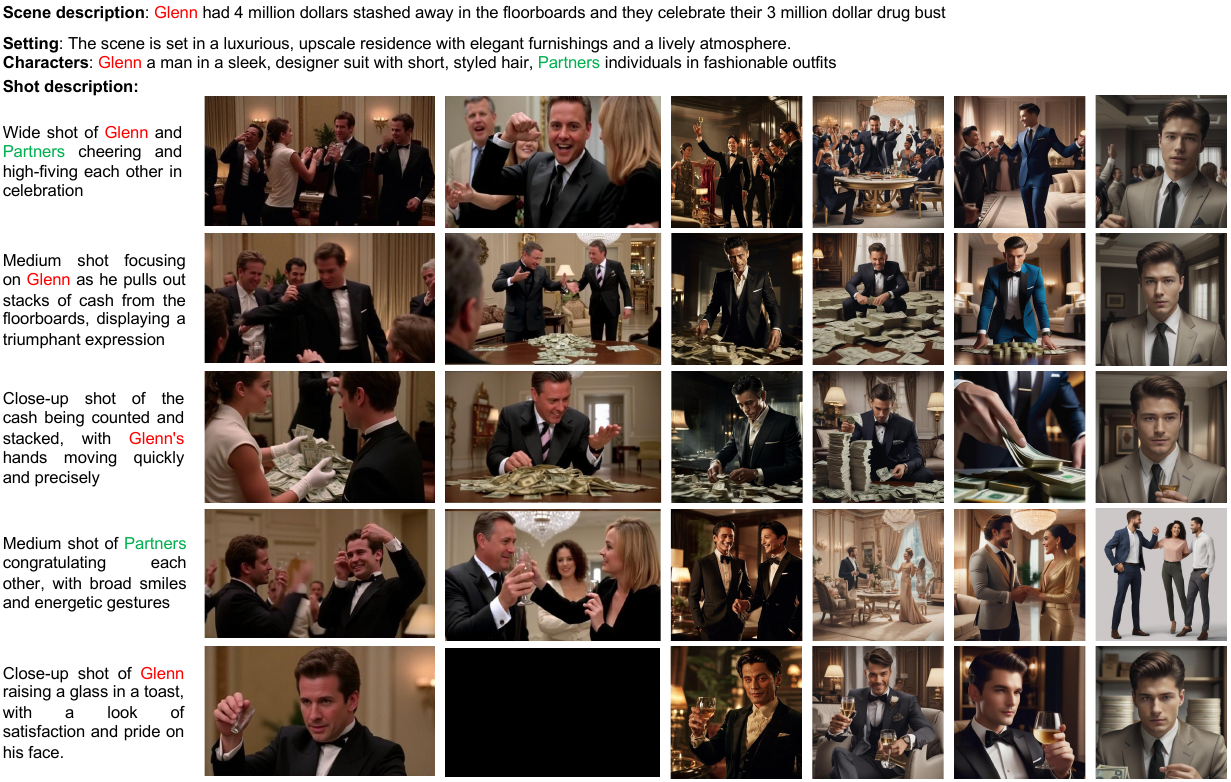}
        \vskip -0.2cm
        \begin{tabu} to 1\textwidth {X[2,c]X[2.2,c]X[2,c]X[1.2,c]X[1.2,c]X[1.2,c]X[1.3,c]}
        &\footnotesize{Ours} & \footnotesize{IC-LoRA~\cite{ic-lora}} &\hspace{0.1cm} \footnotesize{1P1S}~\cite{1prompt}& \footnotesize{ConsiStory}~\cite{consistory} &\footnotesize{StoryDiff}~\cite{storydiff}
        &\footnotesize{VideoStudio}~\cite{videodrafter}
    \end{tabu}
    \end{subfigure}
    
    \vspace{-0.2cm}
    \caption{
    \tb{Visual comparisons with state-of-the-art multi-shot image generation.}
Compared to existing methods, our approach synthesizes coherent keyframes with improved prompt alignment, character consistency, and adherence to specified shot size.
    }
    \label{fig:baseline}
    \vspace{-1em}
\end{figure*}

\noindent\textbf{Baselines}. We compare \name~against five existing state-of-the-art story-to-image generation methods for the cinematic scene composition task: 1) 1Prompt1Story (1P1S)~\cite{1prompt}, 2) ConsiStory~\cite{consistory}, 3) StoryDiff~\cite{storydiff}, 4) VideoStudio~\cite{videodrafter}, and 5) IC-LoRA~\cite{ic-lora}. While the first three achieve character consistency through the self-attention sharing technique, VideoStudio fine-tunes a text-to-image model to generate consistent backgrounds and characters given a background and a foreground image (currently, this method only work with one 
character). IC-LoRA uses a small image set to fine-tune the FLUX model with LoRA to generate a sequence of images at once. The first four are based on UNet~\cite{ronneberger2015u} diffusion models~\cite{ho2020denoising}, and the last uses a DiT~\cite{peebles2023scalable} model. All baselines are configured according to the instructions provided in their respective repositories.

\begin{table}[t]
\centering
\setlength{\tabcolsep}{3pt}
\small
\caption{Quantitative evaluation using CLIP text-image alignment, and DreamSim for different LLM prompting techniques for scene planning. The rightmost column reports the \% of human responses that favor our proposed instruction prompting over the baseline method.}
\label{tab:llm}
\begin{tabu} to 1.0\linewidth {X[1.5,l]X[0.7,c]X[0.7,c]X[2,c]}
\toprule
{Method} & CLIP~$\uparrow$ & DS~$\downarrow$ & Prefered by Human (\%) \\ 
\toprule
Generic prompt  & 0.2050 & \textbf{0.4377} & 73.63 \\ 
Instruction &0.2051  & 0.4700 & 67.23\\ 
Ours &\textbf{ 0.2200} & 0.4472 & -\\ 

\bottomrule
\end{tabu}
\vspace{-2em}
\label{tab:llm}
\end{table}

\subsection{Evaluation metrics}
\label{sec:exp_detail}

In this section, we describe the various metrics we use to evaluate the performance of each approach. We randomly select 800 scene descriptions from different movies for testing, ensuring a diverse and representative test.

\noindent\textbf{Automatic metrics}. Following prior work~\cite{consistory, 1prompt}, we evaluate our generated images using two established metrics: CLIP score and DreamSim (DS) score. The CLIP score~\cite{clip} measures text-image alignment, whereas the DS score ~\cite{dreamsim} measures the consistency of a scene

We also use GPT-4 \cite{gpt4} for evaluating the cinematic scene composition task following the prior work~\cite{makeanything}. Additional details can be found in the App. \ref{sec:metric}.

\noindent\textbf{User studies}. Automated metrics often favor minor layout and appearance variations, which suit traditional story or video generation tasks. In contrast, cinematic scene composition demands consistent portrayal of characters and backgrounds, diverse camera angles capturing key events, and coherent frame sequences to build a compelling narrative. To address these nuances, we conduct an extensive user study to evaluate our generated keyframe sequences.

\begin{table*}[ht]
\vspace{-1.5em}
\centering
\small
\caption{We ask GPT-4 and human participants whether they prefer \name~or a baseline for textual alignment, consistency, and continuity. The values can be interpreted as the \% of responses that favor our method over the baseline.}
\label{tab:comparison}
\begin{tabu} to 0.98\textwidth {@{}X[2.2,l]X[0.7,c]X[1.3,c]X[1.1,c]X[1.3,c]X[1.3,c]X[1.3,c]X[1.3,c]X[1.3,c]X[1.3,c]X[1.3,c]@{}}
\toprule
 & \multicolumn{6}{c}{GPT-4} & \multicolumn{4}{c}{User study}\\
 \cmidrule(lr){2-7} \cmidrule(lr){8-11} 
 & \multicolumn{2}{c}{Textual Align.} & \multicolumn{2}{c}{Consistency} & \multicolumn{2}{c}{Continuity}& \multicolumn{2}{c}{Textual Align.} & \multicolumn{2}{c}{Consistency} \\
\cmidrule(lr){2-3} \cmidrule(lr){4-5} \cmidrule(lr){6-7} \cmidrule(lr){8-9} \cmidrule(lr){10-11}
Ours vs. & Scene & Shot  & Char  & BG  & Action & Camera  & Scene & Shot  & Char  & BG \\
\midrule
1P1S~\cite{1prompt} &  82.29 & 81.25 & 75.00 & 77.08 & 81.25 & 83.33 & 65.56 & 62.96 & 58.42 & 52.00  \\
ConsiStory~\cite{consistory} & 75.51 & 67.35 & 60.20 & 68.37 & 67.35 & 76.53  &66.67 & 52.44 & 53.27 & 64.29 \\
StoryDiff~\cite{storydiff} & 81.11 & 76.67 & 66.67 & 70.00 & 76.67 & 81.11 & 65.38 & 54.07 & 50.00 & 50.00  \\
IC-LoRA~\cite{ic-lora} & 84.41 & 76.62 & 70.12 & 79.22 & 77.92 & 83.11 & 69.10 & 56.82 &66.67 & 52.63  \\
VideoStudio~\cite{videodrafter} & 79.93 &65.61  & 75.59 & 62.28 & 69.00 & 73.00 & 69.26 &  65.93  & 51.92 &61.54  \\

\bottomrule
\end{tabu}
\vspace{-1.5em}
\label{tab:llm_baseine}
\end{table*}

We compare our results against each baseline using a standard two-alternative forced-choice format. Specific details about user study setup can be found in App. \ref{sec:metric}. For each baseline method, we evaluate 200 randomly selected scenes, each containing 3–10 shots. Our study consists of four surveys, each focusing on one key aspect:
\textit{(1)~Scene alignment} assesses how well the generated sequence depicts the progression and continuity of events, characters, setting, and overall storyline described in the scene description;  \textit{(2)~Shot alignment} assesses how well each individual shot depicts its corresponding text description by evaluating the characters' actions, expressions, and shot size; \textit{(3)~Character consistency} assesses whether the main characters' appearances remain uniform across all shots, evaluating factors such as clothing, hairstyle, facial features, and overall identity, even as their actions and expressions vary; \textit {(4)~Setting consistency} evaluates whether the overall setting remains consistent across all frames, even if the exact background or perspective changes. We collect 50 responses per (\name, baseline) pair for each survey, resulting in $\sim$1K responses.

\vspace{-0.1pt}
\subsection{Quantitative results}

\Tref{tab:llm} compares various prompting strategies using CLIP and DS scores, along with user study results on scene alignment. The results show that with appropriate instructions and exemplars, the model adapts well to cinematic planning. Our approach, which combines detailed instructions and exemplars, outperforms both generic prompts and instructions-only methods. \Tref{tab:llm_gpt4} comparing our method with various GPT-4 prompting strategies, we achieve superior results in every aspect. This supports the effectiveness of combining detailed instructions with examples. 

We compute the CLIP and DS scores for each generated image, and report the results separately for scenes containing one, two, and three characters in \Tref{tab:baseline} to enable a thorough comparison between our approach and the baselines.
Our method outperforms all baselines in textual alignment by following complex guidance on camera shots, character appearance, actions, and expressions. We note that some misalignments still occur due to the challenges of text-to-image generation models with handling long input prompts. While baseline methods show slightly higher DS scores due to the metric's bias toward static perspectives, our approach deliberately generates varied angles to enhance scene storytelling. 

The user study and GPT-4 evaluation in \Tref{tab:llm_baseine} shows a preference for our method over the baselines across all metrics. Although some automated metrics may favor the baselines, human judgments and GPT-4 assessments, which are more reliable for evaluating the cinematic scene composition task, demonstrate that our approach performs better.
Moreover, we compare our approach to IC-LoRA for 3 to 10 frames, and overall, our method significantly improves accuracy. For instance, at three frames, our method achieves 95.45\% accuracy compared to 34.84\% for IC-LoRA, a trend that continues across higher frame counts, demonstrating robust and superior performance.

\begin{table}[t]
\centering
\setlength{\tabcolsep}{3pt}
\small
\caption{Accuracy between our text-to-image and IC-LoRA. with different number of shots.}
\label{tab:number}
\begin{tabu} to 0.48\textwidth {X[2.2,l]X[1.1,c]X[1.1,c]X[1.1,c]X[1.1,c]X[1.1,c]X[1.1,c]X[1.1,c]X[1.1,c]}
\toprule
{$\#$ shots} & {3}  &{4} & {5} & {6} & {7} & {8} & {9} & {10}  \\ 
\midrule
IC LoRA & 34.84 & 37.60 & 17.64 & 16.67 & 26.08 &  21.05 & 08.33 & 16.66 \\
Ours &   95.45 & 91.45 & 91.17 & 85.98 & 78.87  & 72.72 & 60.24 & 42.24\\ 
\bottomrule
\end{tabu}

\end{table}

\vspace{-0.1pt}
\subsection{Qualitative results}
\Fref{fig:baseline} presents qualitative comparisons between our method and the baselines. VideoStudio~\cite{videodrafter} produces nearly identical frames that poorly follow text instructions, StoryDiffusion~\cite{storydiff} suffers from inconsistent character identities, and 1P1S~\cite{1prompt} shows fluctuating background themes. Although IC-LoRA~\cite{ic-lora} maintains good continuity and character consistency, it struggles with text alignment, shot size accuracy, and is sometimes missing frames. In contrast, our method achieves better text alignment, accurate shot sizes, and the correct frame count, resulting in a more coherent scene. 

\subsection{Ablation studies}

Our ablation studies systematically investigated the impact of four key hyperparameters on model performance, as shown in \Tref{tab:ablation}: LoRA rank, training iteration count, addition of borders between training shots, and scene/shot balancing. Among LoRA configurations, a rank of 128 yielded the highest accuracy (88.83\%) and best CLIP score (0.2118), indicating optimal capacity for feature adaptation. Analyzing the number of training iterations, we observed that performance improved steadily as iterations increased up to 16k steps, achieving peak accuracy (88.83\%), but notably declined at 20k steps, suggesting potential overfitting. Including explicit borders between training shots significantly enhanced accuracy (88.83\% vs. 47.20\%) and improved CLIP scores (0.2121 vs. 0.1960), highlighting the importance of clear visual delineation. Lastly, training on balanced scene/shot data further boosted accuracy to 88.83\% , emphasizing that balanced training data is crucial for accurate shot-count generation. Consequently, the optimal hyperparameter combination identified was LoRA rank 128, training for 16k iterations, inclusion of borders between shots, and using balanced data.

 \begin{table}[t]
\setlength{\tabcolsep}{3pt}
\small
\centering
\caption{Ablation studies on different training settings.}
\label{tab:llm}
\begin{tabu} to 0.45\textwidth {X[0.5,l]X[1.2,c]X[1.2,c]X[1.2,c]}
\toprule
 &  {Acc.(\%)} $\uparrow$ & {CLIP }$\uparrow$ & {DS}$\downarrow$ \\ 
\midrule
\multicolumn{4}{l}{\textit{\textbf{LoRA rank/alpha}}} \\
32 & 73.10   & 0.2048 & \textbf{0.4145}\\ 
64 & 77.83   &  0.2073 & 0.4539\\
128 & \textbf{88.83} & \textbf{0.2118} & 0.4476\\ 
\bottomrule
\multicolumn{4}{l}{\textit{\textbf{Training iteration}}}\\
2k & 75.24 & 0.2085 & 0.3853 \\
5k & 77.15 &  0.2068 & 0.3998 \\
10k&  79.03 &\textbf{0.2165}& 0.5501\\ 
15k & 87.58 & 0.2099 & 0.4499\\ 
16k &  \textbf{88.83} & 0.2118 & 0.4476 \\ 
20k &  72.42 & 0.2018 & \textbf{0.3476} \\ 
\bottomrule
\multicolumn{4}{l}{\textit{\textbf{Adding border between training shots}}} \\
\xmark & 47.20 &  0.1960 & \textbf{0.4322}\\
\cmark &  \textbf{88.83}  & \textbf{0.2121} & 0.4476 \\ 
\bottomrule
\multicolumn{4}{l}{\textit{\textbf{Scene/shot balancing}}}  \\
\xmark & 57.86 & 0.1993 &  0.47612 \\
\cmark &    \textbf{88.83} & \textbf{0.2118} & \textbf{0.4476} \\ 
\bottomrule
\end{tabu}
\label{tab:ablation}
\end{table}

\subsection{Limitation and discussion}
\begin{figure}[t]
  \centering
    \includegraphics[width=1.0\columnwidth]{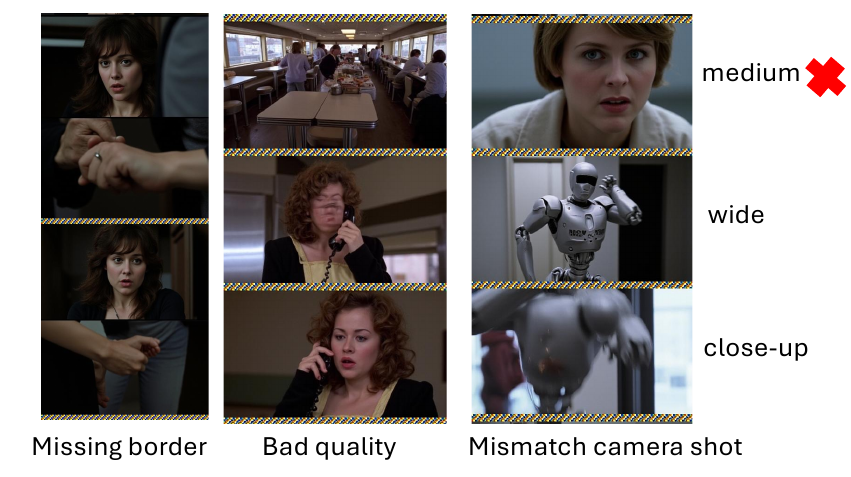}
  \caption{
  \tb{Limitation.} Our method sometimes still suffers from bad image quality with artifacts, such as missing borders, and mismatches with the shot size specified in the shot description.
.
}
  \label{fig:limitation}
\end{figure}

We have improved textual alignment and image separation; however, our method still suffers from artifacts, missing borders, and occasional mismatches with the text prompt, as depicted in ~\Fref{fig:limitation}. We plan to address these issues in future work. 

\begin{table}[t]
\centering
\setlength{\tabcolsep}{3pt}
\small
\caption{Quantitative evaluation using CLIP text-image alignment (CLIP), and DreamSim (DS).}
\label{tab:baseline}
\begin{tabu} to 0.49\textwidth {@{}X[3.0,l]X[1.3,c]X[1.3,c]X[1.1,c]X[1.3,c]X[1.3,c]X[1.3,c]@{}}
\toprule
 \multirow{2}{*}{Method}& \multicolumn{2}{c}{1 Char.} & \multicolumn{2}{c}{2 Char.} & \multicolumn{2}{c}{3 Char.} \\
\cmidrule(lr){2-3} \cmidrule(lr){4-5} \cmidrule(lr){6-7} 
& CLIP~$\uparrow$ & DS $\downarrow$ & CLIP~$\uparrow$ & DS $\downarrow$ & CLIP~$\uparrow$ & DS $\downarrow$ \\
\midrule
1P1S~\cite{1prompt} & 0.2213 &0.2677  & 0.2067 & \textbf{0.2444}  & 0.1954 & \textbf{0.2141} \\
ConsiStory~\cite{consistory} & 0.2085  & 0.3497  & 0.2022 & 0.3772  & 0.1985 & 0.3932\\
StoryDiff~\cite{storydiff} &0.2021  &0.3915    & 0.1945 &0.4147 & 0.1867 &0.4252 \\
VideoStudio~\cite{videodrafter} & 0.2173 & \textbf{0.2635} & - & - & -&-  \\
IC-LoRA~\cite{ic-lora} & 0.2051 &0.3630 & 0.1935 &0.3922 & 0.1897 &0.4121 \\
Ours   & \textbf{0.2236} & 0.5482 & \textbf{0.2095}  & 0.5541 & \textbf{0.2044} &  0.5610   \\
\bottomrule
\end{tabu}
\vspace{-1em}
\end{table}

\label{sec:exp_quan}

\vspace{-1pt}
\section{Conclusion}
\label{sec:conclucion}
We present \name, a novel two-stage framework for cinematic scene composition that leverages LLM-based planning and a fine-tuned text-to-image model. By first generating a detailed plan from high-level descriptions and then synthesizing high-quality keyframes, our approach addresses the challenges of maintaining consistency and continuity, handling multiple characters, and capturing cinematic effects. Experimental results demonstrate that \name~produces contextually-rich keyframes to create a coherent visual narrative.

{\small
\bibliographystyle{ieeenat_fullname}
\bibliography{main}
}
\clearpage
\appendix

\section{Appendix}
\label{sec:appendix}

\subsection{Additional qualitative results}
\label{sec:append_qualitative}
\begin{figure*}[h]
    \begin{subfigure}[b]{1.\textwidth}
        \centering
        \includegraphics[width=1.\textwidth]{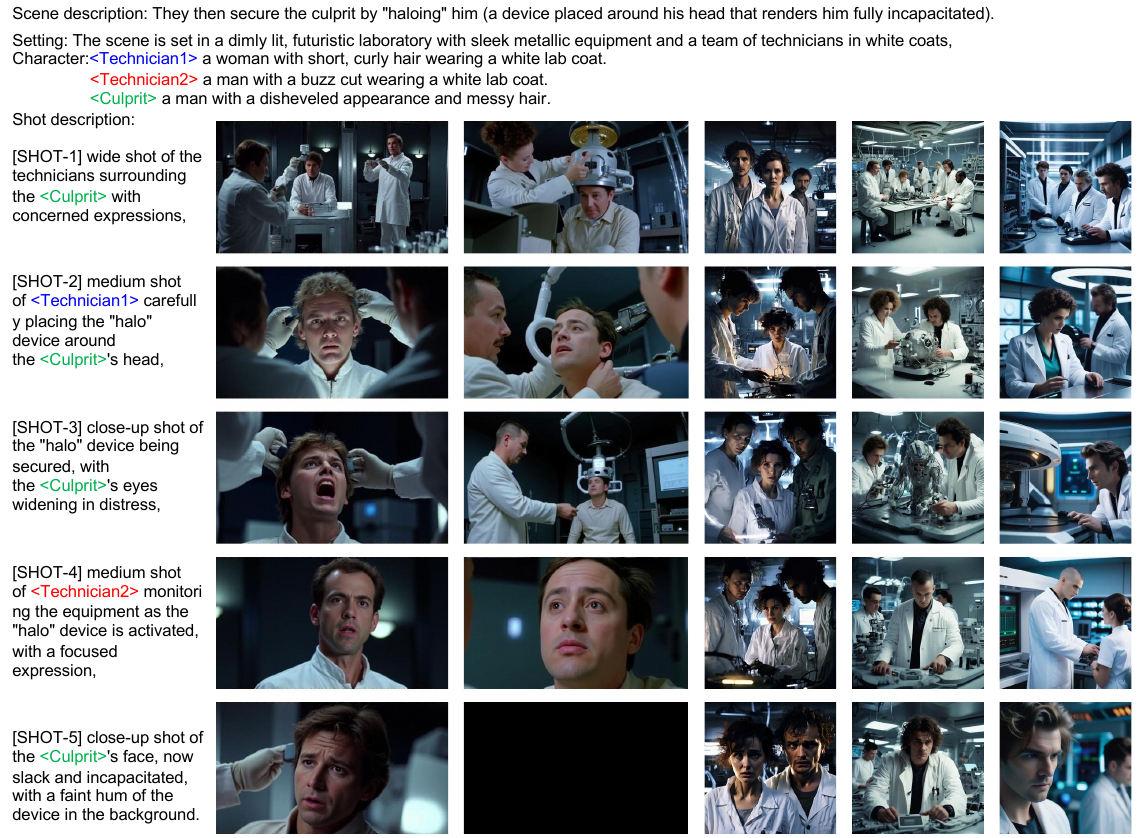}
        \vskip -0.1cm
        \begin{tabu} to 1\textwidth {X[2,c]X[2.2,c]X[2,c]X[1.2,c]X[1.2,c]X[1.2,c]}
        &\footnotesize{Ours} & \footnotesize{IC-LoRA~\cite{ic-lora}} &\hspace{0.01cm} \footnotesize{1P1S}~\cite{1prompt}& \footnotesize{ConsiStory}~\cite{consistory} &\footnotesize{StoryDiff}~\cite{storydiff}
        
    \end{tabu}
    \end{subfigure}
    
    \vspace{-0.2cm}
    \caption{
    \tb{Additional visual comparisons with state-of-the-art multi-shot image generation.}
Our approach generates coherent keyframes with superior prompt alignment, consistent characters, and precise shot sizing.
    }
    \label{fig:app1}
    \vspace{-1em}
\end{figure*}

\begin{figure*}[h]
    \begin{subfigure}[b]{1.\textwidth}
        \centering
        \includegraphics[width=1.\textwidth]{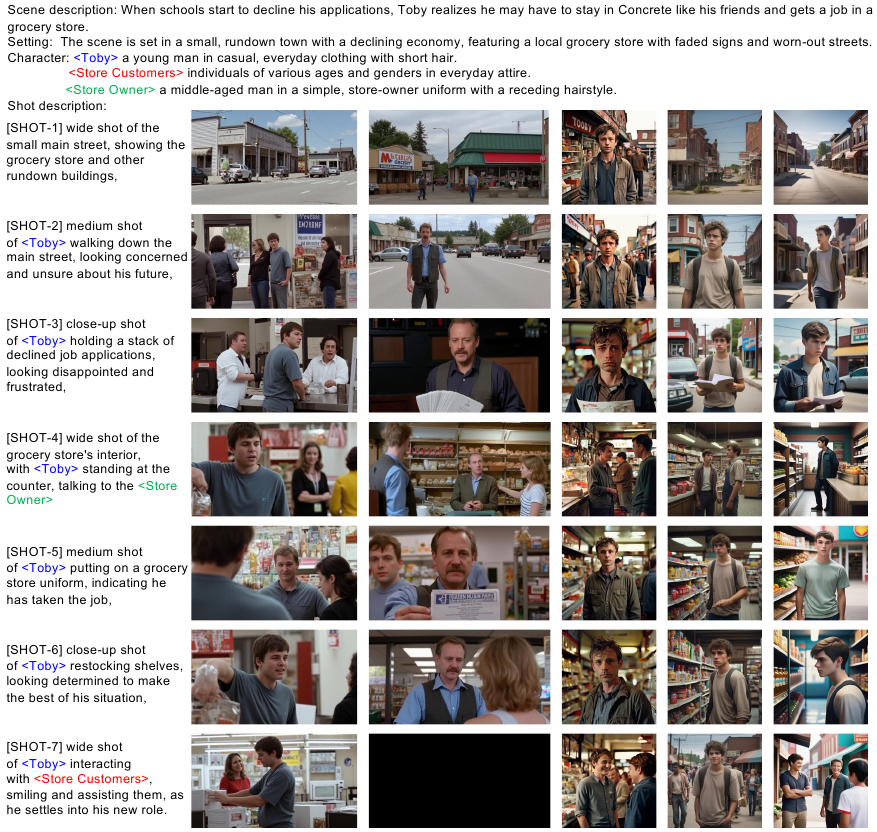}
        \vskip -0.1cm
        \begin{tabu} to 1\textwidth {X[2,c]X[2.2,c]X[2,c]X[1.2,c]X[1.2,c]X[1.2,c]}
        &\footnotesize{Ours} & \footnotesize{IC-LoRA~\cite{ic-lora}} &\hspace{0.01cm} \footnotesize{1P1S}~\cite{1prompt}& \footnotesize{ConsiStory}~\cite{consistory} &\footnotesize{StoryDiff}~\cite{storydiff}
        
    \end{tabu}
    \end{subfigure}
    
    \vspace{-0.2cm}
    \caption{
    \tb{Additional visual comparisons with state-of-the-art multi-shot image generation, showcasing an increased number of shots)}
Our approach performs well even with a larger number of shots, maintaining character consistency and narrative coherence.
    }
    \label{fig:app2}
    \vspace{-1em}
\end{figure*}

We include additional qualitative comparisons of our method against the baselines shown in  \Fref{fig:app1} and \Fref{fig:app2}. Our approach generates coherent keyframes with superior prompt alignment, consistent character portrayal, and precise shot sizing. It continues to perform robustly even as the number of shots increases, preserving both character consistency and overall narrative coherence. 

\begin{figure}[!ht]    
    \begin{subfigure}[b]{0.49\textwidth}
        \centering
        \includegraphics[width=1\textwidth]{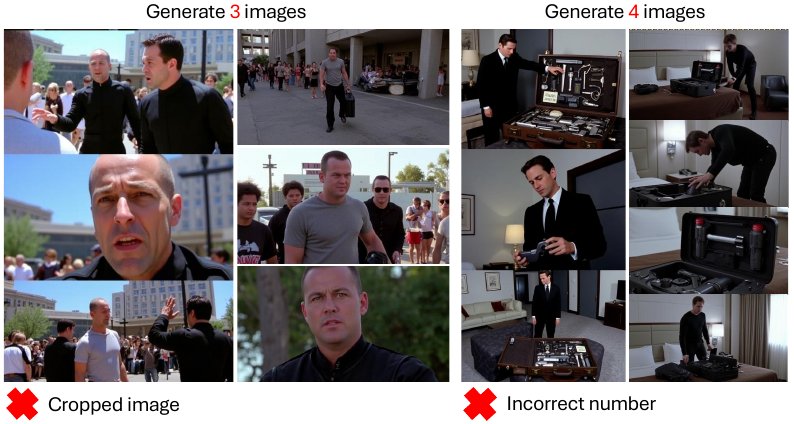}
        \vskip -0.1cm
        \begin{tabu} to 1\textwidth {X[2.2,c]X[2.2,c]X[1.7,c]X[1.5,c]}
        \footnotesize{IC-LoRA} & \footnotesize{Ours} & \footnotesize{IC-LoRA}& \footnotesize{Ours} 
    \end{tabu}
    \end{subfigure}

    \vspace{-0.2cm}
    \caption{
    \tb{The problem of IC-LoRA }.
    IC-LoRA often generates cropped frames and/or the incorrect number of images.
    }
    \label{fig:lora}
    \vspace{-1em}
\end{figure}

\Fref{fig:lora} illustrates some of the issues with IC-LoRA ~\cite{ic-lora}, such as generating cropped images or misalignment with the input prompt. These problems are due to their data processing pipeline.

\subsection{Dataset discussion}
\label{sec:append_data}
\begin{figure*}[t]
    \centering
    \vspace{-2em}
    \includegraphics[width=1.0\textwidth]{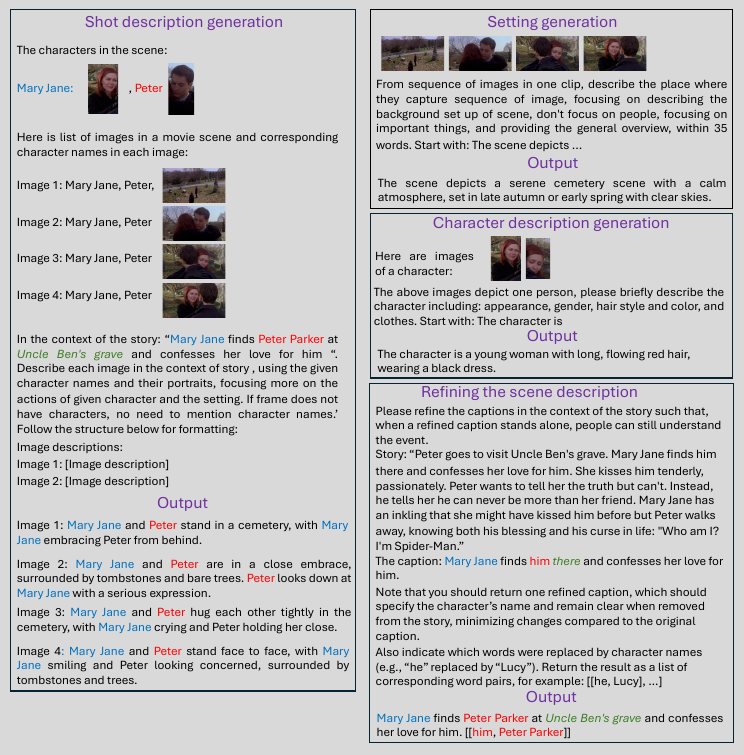}
    \caption{
    \textbf{Instruction to extract data attributes using LLaVa-OneVision}.
We provide detailed instruction for 4 tasks: shot description, setting, character description generation and refining scene description.
    }
    \label{fig:instruction}
    \vspace{-1em}
\end{figure*}

\textbf{Prompting using for LLaVa-OneVision}. 
To leverage LLaVA-OneVision 72B for our visual understanding task, we provide a straightforward instruction format followed by images. There are three main extracted information: setting, shot description, and character description.  An example instruction prompt for generating the setting, character descriptions, and per-shot descriptions is shown in \Fref{fig:instruction}. Given the input frames and detailed instructions, LLaVA-OneVision demonstrates strong task comprehension. As a result, our new dataset, constructed using MLLMs, provides richer shot-level contextual information than existing datasets, as shown in \Tref{tab:data}.

\begin{table}[H]
\centering
\setlength{\tabcolsep}{3pt}
\small
\caption{Comparison of \name~with other datasets: CondensedMovie, MSA, MovieNet, Storyboard20k. Our dataset is rich with attributes at the shot level.  }
\label{tab:data}
\begin{tabu} to 0.49\textwidth {X[3,l]X[1.1,c]X[1.1,c]X[1.1,c]X[1.1,c]X[1.1,c]X[1.1,c]X[1.1,c]}
\toprule
{Dataset} & $\#$ movie  & $\#$ scene & $\#$ shot&  Shot Desc. & Char. Desc. & Setting & Cam. Shot \\ 
\midrule
\footnotesize{CondensedMovie} & 3.6K & 33K  & 400K & \xmark &\xmark& \xmark &\xmark  \\
\footnotesize{MSA} &  327 & 4.5K & - & \xmark &\xmark& \xmark &\xmark \\
\footnotesize{MovieNet} & 1.1K & 43K & 3.9M  &  \xmark  & \xmark & \xmark & \xmark   \\
\footnotesize{Storyboard20k} & 400 & 20K & 150K &\xmark & \xmark & \xmark & \xmark \\
\name &  312 & 10K &46K  & \cmark  & \cmark  & \cmark  & \cmark  \\
\bottomrule
\end{tabu}

\end{table}

\subsection{Additional evaluation metrics}
\label{sec:metric}
We evaluate the accuracy of generating the correct number of images in the scene, as specified by the input scene plan. A scene is considered ``correct'' if its shot count matches the expected number. For sequences generated by IC-LoRA, where images lack clear borders between shots, we estimate transitions between frames by calculating the pixel difference between adjacent rows, with the highest difference indicating the separation boundary. Since we incorporate distinct borders to separate the frames for training \name, we utilize the Canny edge detector to identify these borders. We show a comparison of our method to IC-LoRA in \Tref{tab:number}.

\textbf{MLLM evaluations}. We leverage LLaVA-OneVision, a state-of-the-art MLLM capable of understanding image sequences through visual narrative analysis, for evaluations. Additionally, prior work has demonstrated that GPT-4 can evaluate image sequences with strong alignment to human judgment. Thus, we use both GPT-4 and LLaVA-OneVision to assess the performance of our method for the cinematic scene composition task. As shown in \Tref{tab:llava1}, our method consistently outperforms all baselines across evaluation metrics, with similar trends observed in both GPT-4 and human evaluations.

Since MLLMs can assess multiple aspects of a visual sequence, we expand our evaluation beyond the initial four criteria to include two additional ones:
\begin{enumerate}
    \item \textbf{Action flow}: Assess whether the sequence displays a smooth and logical progression of actions and expressions that reflect the scene's dynamics.
    \item \textbf{Camera movement}: Determine whether transitions between keyframes resemble coherent, movie-like camera motions that enhance storytelling.
\end{enumerate} 
We evaluate 200 images per baseline. Following the structure of the user study, each question compares our method against a single baseline. To mitigate biases and ensure fairness, we provide clear instructions to guide the MLLM in selecting the better image sequence for each criterion and randomize the side on which each method appears for all trials. The full instruction prompt is shown in \Fref{fig:metrics}.

\begin{table}[ht]
\centering

\small
\caption{Comparison of our method and the baselines using LLaVA-OneVision. }
\label{tab:llava1}
\begin{tabu} to 0.98\linewidth {@{}X[2,l]X[1.0,c]X[1.0,c]X[1.0,c]X[1.0,c]X[1.0,c]X[1.3,c]@{}}
\toprule
 & \multicolumn{2}{c}{Textual Align.} & \multicolumn{2}{c}{Consistency} & \multicolumn{2}{c}{Continuity}\\
\cmidrule(lr){2-3} \cmidrule(lr){4-5} \cmidrule(lr){6-7}
Ours vs. & Scene & Shot  & Char  & BG  & Action & Camera\\
\midrule
1P1S &   82.34 & 82.32 & 81.43 & 82.34 & 83.52 & 82.43 \\
ConsiStory & 65.45 & 63.63 & 65.45 & 63.63  & 65.45 & 65.45 \\
StoryDiff & 82.43 & 83.42 & 82.43 & 84.56 & 83.45 & 83.45  \\
IC-LoRA & 73.23 & 71.14 & 74.22 & 70.21 & 69.53 & 74.23 \\

\bottomrule
\end{tabu}
\end{table}

\begin{figure*}[t]
    \centering
    \vspace{-2em}
    \includegraphics[width=1.0\textwidth]{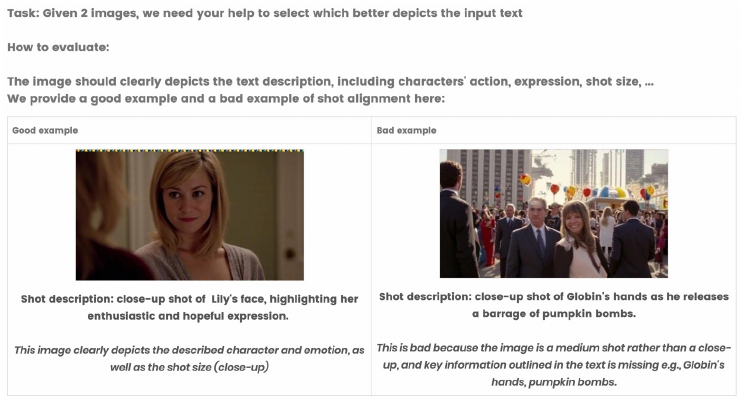}
    \caption{
    \textbf{User study}.
The instructions presented to users at the beginning of the survey.
    }
    \label{fig:userstudy}
    \vspace{-1em}
\end{figure*}

\textbf{User study}. Participants first receive detailed instructions at the beginning of the survey, including task explanations and illustrative examples distinguishing good from bad cases (an example can be found \Fref{fig:userstudy}). These initial instructions ensure better comprehension of the survey aspects. Additionally, each question in the survey has a time limit, helping to reduce noise by preventing users from lingering excessively on difficult-to-decide questions. 

\subsection{Prompting LLMs for scene planning and evaluations }
\label{sec:gpt-4}

\begin{figure*}[t]
    \centering
    \includegraphics[width=1\textwidth]{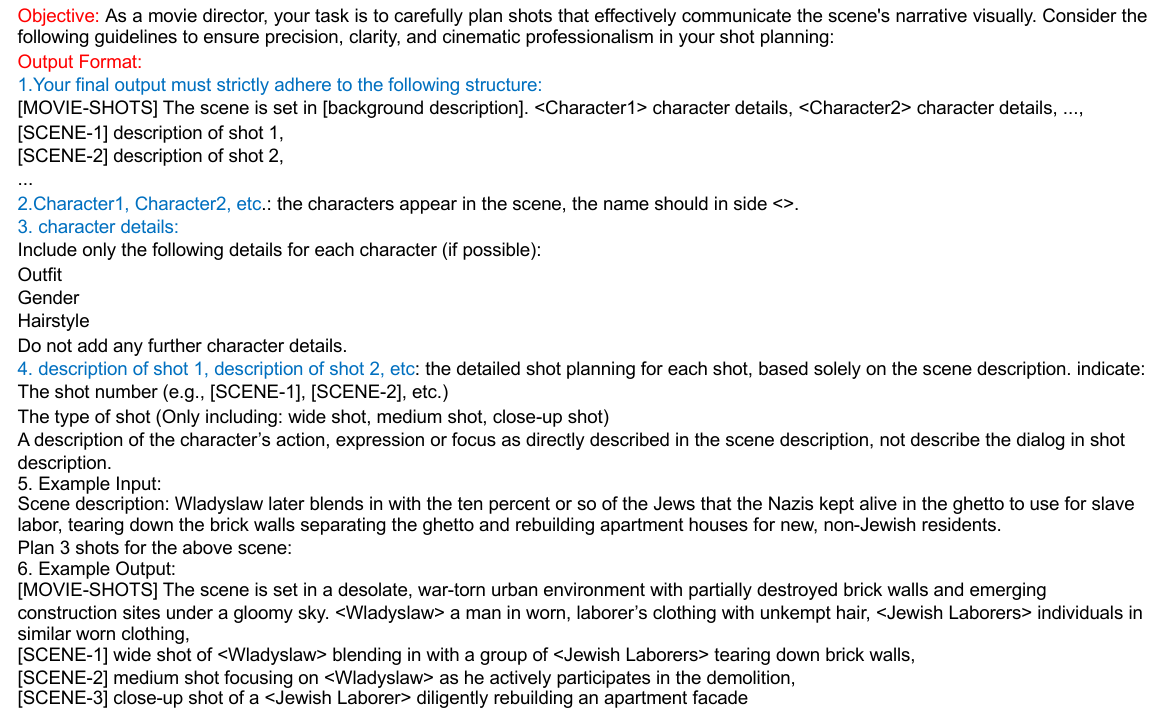}
    \caption{
    \textbf{Scene planning instruction prompt}. Example of a prompt used to guide LLMs in the \textit{scene planning} stage of \name.
    }
    \label{fig:prompt}
\end{figure*}

\Fref{fig:prompt} shows the full instructions used in the shot planning stage, guiding the LLM to produce outputs with the correct format and accurate attributes for generation. \Fref{fig:metrics} presents the complete evaluation prompts used by GPT-4 and LLaVA-OneVision to assess the consistency and quality of generated image sequences within the same scene.

\begin{figure*}[h]
\centering

\begin{minipage}{\textwidth}
\lstset{
  basicstyle=\ttfamily,
  breaklines=true,
  frame=single,
  backgroundcolor=\color{white},
  showstringspaces=false
}
\begin{lstlisting}


You are an expert in movie scene analysis. You will be given 2 sequence of images, the two representing one scene from a movie. Your job is to evaluate and select the better sequence that best exemplifies the scene based on three specific criteria.
    1. Textual Alignment:
        * Overall Scene: Assess how well the keyframes capture the narrative, mood, and setting as described in the overall scene description.
        * Shot Details: Evaluate how accurately each keyframe reflects the detailed descriptions provided for individual shots.
        * Key Points: Consider whether the depicted actions, expressions, and visual details align with both the story and shot specifics
    2. Consistency:
        * Character Consistency: Ensure that the main character's appearance (clothing, hairstyle, facial features) remains uniform across all keyframes, even as their actions vary.
        * Background Consistency: Verify that the backgrounds, although possibly shown from different perspectives, clearly indicate the same location.
    3. Continuity:
        * Action Flow: Analyze the sequence for smooth and logical progression of actions and expressions that mirror the described scene's dynamics.
        * Camera Movement: Evaluate if the camera transitions and shifts between keyframes create a coherent, movie-like progression that enhances the storytelling
    For each answer, you should explain why you choose this option. 
    Then the final answer should be the choosen sequence (the best sequence) for each aspect like bellow format, the choosen sequence can be different for different aspects:
    1. Textual Alignment: 
        * Overall Scene: [chosen sequence]
        * Shot Details:[chosen sequence]
        * Key Points: [chosen sequence]
    2. Consistency: 
        * Character Consistency: [chosen sequence]
        * Background Consistency: [chosen sequence]
    3. Continuity: 
        * Action Flow: [chosen sequence]
        * Camera Movement: [chosen sequence]
    

\end{lstlisting}
\end{minipage}
\caption{\textbf{Evaluation instruction prompt.} Instructions for  GPT-4 and LLaVA-OneVision to assess the results of the keyframe generation stage.}
\label{fig:metrics}
\end{figure*}

\end{document}